\documentclass{article}
\usepackage{ijcai26}
\usepackage{tcolorbox}
\usepackage{times}
\usepackage{soul}
\usepackage{url}
\usepackage[hidelinks]{hyperref}
\usepackage[utf8]{inputenc}
\usepackage[small]{caption}
\usepackage{graphicx}
\usepackage{amsmath}
\usepackage{amsthm}
\usepackage{amssymb}
\usepackage{booktabs}
\usepackage{algorithm}
\usepackage{algpseudocode}

\usepackage{algcompatible}
\usepackage[switch]{lineno}
\usepackage[dvipsnames]{xcolor}
\usepackage{xurl}

\title{Improving Multimodal Reasoning via Worst Dimension Optimization}

\author{
Haocheng Lv
\and
Huaping Zhang
\and
Qiuchi Li
\and
Lei Li
\And
Chunxiao Gao\thanks{Corresponding author.}\\
\affiliations
Beijing Institute of Technology\\
\emails
\{3120255822, kevinzhang, liqiuchi,lilei,gao\_chunxiao\}@bit.edu.cn}

\begin{document}

\maketitle

\begin{abstract}
Multimodal reasoning requires a path that retains integrity over a wide range of constraints, from visual grounding to logic consistency. However, the current Process Reward Models (PRMs) focus on heuristically defined rewards that equally weigh these factors, which may lead to the concealment of individual dimension failures (such as visual hallucinations) by the dominating factors, without guaranteeing the validity of the reasoning process in general. Therefore, to overcome the limitation, the paper proposes the concept of Multimodal Multi-Dimensional Scalarization Process Reward Modeling (MMS-PRM), a paradigm specifically developed to enforce the worst dimension’s robustness in multimodal reasoning. Specifically, a hierarchical fine-grained reward space is developed to represent the multimodal risks in the reasoning tasks, and a Chebyshev-based Monte Carlo Tree Search (MCTS) algorithm is introduced, in which the primary focus during the path searching is given to the worst-performing dimension. Moreover, a curriculum-based Direct Preference Optimization (DPO) approach is developed to gradually learn the balanced reasoning skills in the policy. The experimental results show that, without the dimension collapse issue, the MMS-PRM approach significantly improves the reliability of the multimodal reasoning performance and reaches competitive results in various challenging tasks. The code is available at \url{https://github.com/leibniz-Man/MMS-PRM}.

\end{abstract}

\section{Introduction}

\begin{figure}[!t]
\centering
\includegraphics[width=\linewidth]{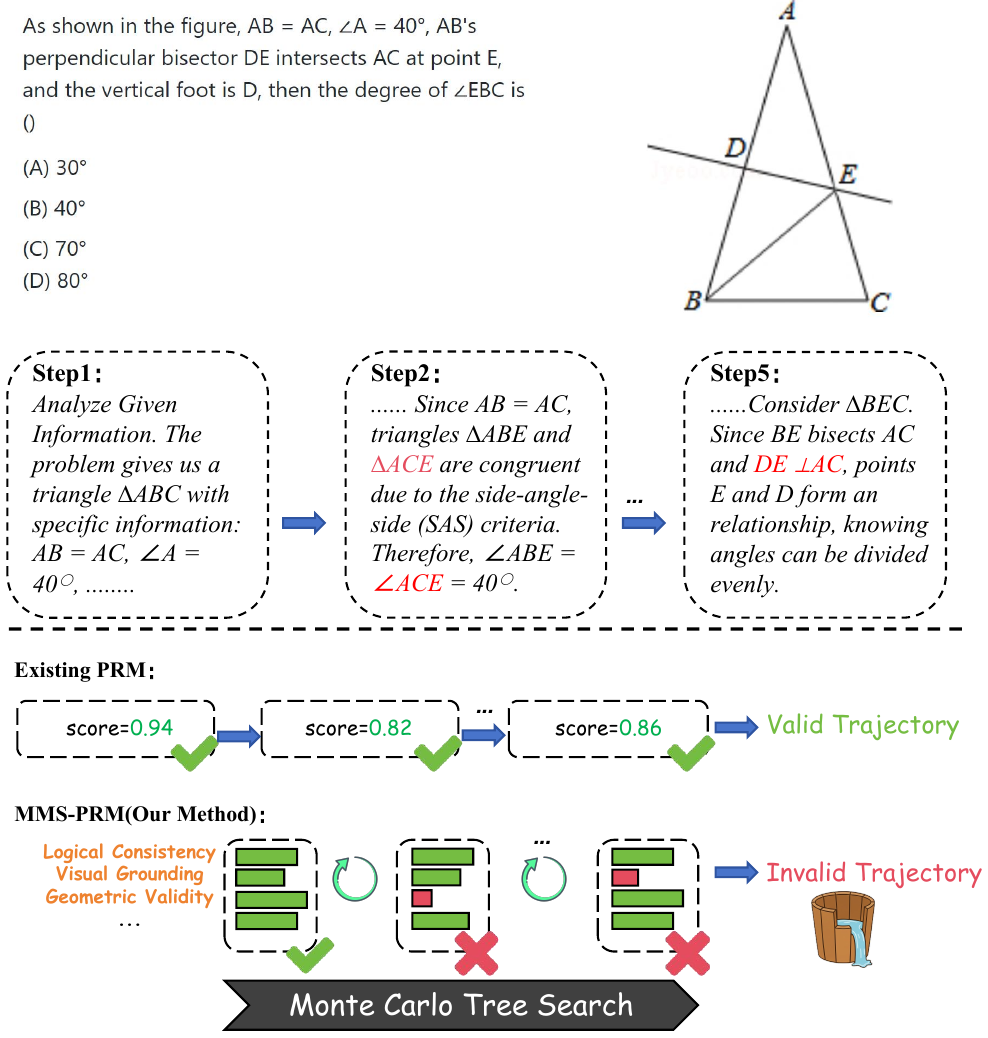}
\caption{Comparison Between Existing Methods and Our Method.}
\end{figure}
Multimodal Large Language Models (MLLMs) are shown to perform well on complex reasoning tasks such as mathematical diagrams and scientific figures. In contrast to the reasoning task on pure-text reasoning, multimodal reasoning requires satisfying multiple constraints simultaneously, including visual grounding and logical correctness, where violation of any single aspect invalidates the entire reasoning trajectory.

To supervise such processes, prior work introduces Process Reward Models (PRMs) that provide step-level feedback. However, existing PRMs~\cite{wang2025visualprm,luo2025ursa,ong2025training,gao2025benchmarking} typically collapse multiple quality dimensions into a single scalar reward. This makes it possible for good performance in some factors to compensate bad performance in other factors, which causes incorrect reasoning paths to be reinforced.

The problem caused by the compensation mechanism is even more severe in multimodal reasoning tasks. As shown in Figure 1, the reasoning process can have a mathematically well-organized reasoning chain with the use of the hallucinated visual relations. Because the text logic is well-organized, the reasoning process will normally be assigned with a high confidence score by the scalar PRM, without punishing the reasoning process for the factual error. The failure indicates that there is a severe weakness in the current averaging approach for designing the reward.

Such findings lead to the primary tenet in this paper: \emph{the integrity of an active multimodal reasoning trajectory is measured not by average quality but by the worst active dimension.} An active reasoning step that is valid but unground in images is inherently invalid and should not be reinforced. Therefore, successful multimodal alignment should progress from optimizing for expectation maximization rewards to non-compensatory goals that punish worst-dimension degeneration.

To operationalize this principle, we propose MMS-PRM, a search-enhanced, multi-dimensional process reward framework for multimodal reasoning. We reformulate multimodal process rewarding as a multi-objective trajectory optimization problem, where the quality of each reasoning step and consequently the whole reasoning trajectory is governed by its weakest relevant dimension rather than an aggregated score. First, we construct a hierarchical, fine-grained reward space that decomposes multimodal reasoning quality into interpretable dimensions and sub-dimensions, allowing rewards to be dynamically activated based on the evolving reasoning context. Second, we introduce a Chebyshev-guided Monte Carlo Tree Search (MCTS) that explicitly prioritizes the worst-performing reward dimension during trajectory exploration, preventing compensation across conflicting criteria and promoting balanced reasoning paths. Finally, we integrate a curriculum-style Direct Preference Optimization (DPO) strategy that progressively trains MLLMs with our obtained balanced reasoning trajectories, from short, high-confidence trajectories to long-horizon multimodal reasoning chains.

Through this closed-loop framework, MMS-PRM enforces balance between visual grounding, logical coherence, and semantic correctness at both step and trajectory levels. Extensive experiments on diverse and challenging multimodal reasoning benchmarks demonstrate that MMS-PRM significantly improves reasoning reliability and robustness, particularly on long-horizon and visually demanding tasks.

The contributions of this work are as follows:
\begin{itemize}
\item We identify a fundamental limitation of scalar process rewards in multimodal reasoning and reformulate process reward modeling as a non-compensatory, multi-dimensional trajectory optimization problem.
\item We propose a worst-dimension-aware search framework that combines hierarchical process rewards with Chebyshev-scalarized MCTS to enforce balanced multimodal reasoning.
\item We introduce a curriculum-based preference alignment strategy that effectively transfers search-discovered reasoning behaviors into the model, achieving strong performance and generalization across benchmarks.
\end{itemize}

\section{Related Work}
\textbf{VLM Reasoning.}
As VLMs continue to be used for increasingly complex tasks in mathematics and science\cite{yue2024mmmu,yao2025countllm,chen2025dense,yan20243dsceneeditor,zhang2026psgs,guan2025learning,liu2024graph}, the need for improving their reasoning ability has become crucial. Previous studies aligning visual regions with reasoning steps\cite{shao2024visual,Yan_2026_WACV,jia2026ram,cai2025role,cai2025bayesian}, or decomposing long-chain reasoning via multi-agent frameworks\cite{dong2025insight,shi2026intrinsic,li2026multiple,li2025human,li2024image}. These advances demonstrate the importance of modeling intermediate reasoning steps\cite{zhang2025improve}. Nevertheless, the majority of the previous studies use coarse-grained reasoning supervision\cite{li2024llava,shao2024visual,li2024image}. In contrast, our work introduces a fine-grained, structured reasoning paradigm to more precisely enhance VLM reasoning. \\
\noindent \textbf{Process Reward Model.}
With more complex tasks being addressed by LLMs, there is a need to reason over longer trajectories, rendering it inadequate to check coherence and correctness of reasoning using Outcome reward model(ORM) \cite{snell2024scaling,luo2024improve,cai2025bayesian}. Early PRM work focuses on objective domains such as mathematics, where intermediate steps admit clear correctness criteria, either by directly modeling step-wise correctness \cite{wang2024math} or estimating its likelihood \cite{wang2024multi,guan2025rstar}.
Recent studies extend PRMs to multimodal reasoning. VisualPRM \cite{wang2025visualprm} aligns intermediate reasoning with visual evidence via step-level rewards, URSA \cite{luo2025ursa} integrates process supervision into policy learning through reinforcement learning, while VL-PRM300K \cite{ong2025training} and SVIP \cite{gao2025benchmarking} enable detailed multimodal supervision with large-scale annotated datasets and visual programming. The current state-of-the-art in multimodal PRM is based on scalarized rewards, which may hiding important dimensions. To address this issue, we introduce the MMS-PRM, which represents multimodal process-level supervision in a multi-dimensional reward space to capture the quality of reasoning more accurately. \\
\noindent \textbf{Tree-based Search in LLMs.}
Tree structures have demonstrated significant potential in language models \cite{qi2025mutual,wu2024beyond}. Recent efforts explore applying these tree search methods to identify effective reasoning paths for MLLMs. AR-MCTS \cite{dong2025progressive} enhances multimodal reasoning by integrating MCTS with active retrieval, but its high computational overhead and extensive iterations limit its practicality. Similarly, Mulberry \cite{yao2026mulberry} distills 260K long-chain reasoning samples via tree structures from powerful models like GPT-4o, but relies heavily on resource-intensive teacher models.

\section{Method}
\label{sec:method}
\begin{figure*}[!t]
\centering
\includegraphics[width=\linewidth]{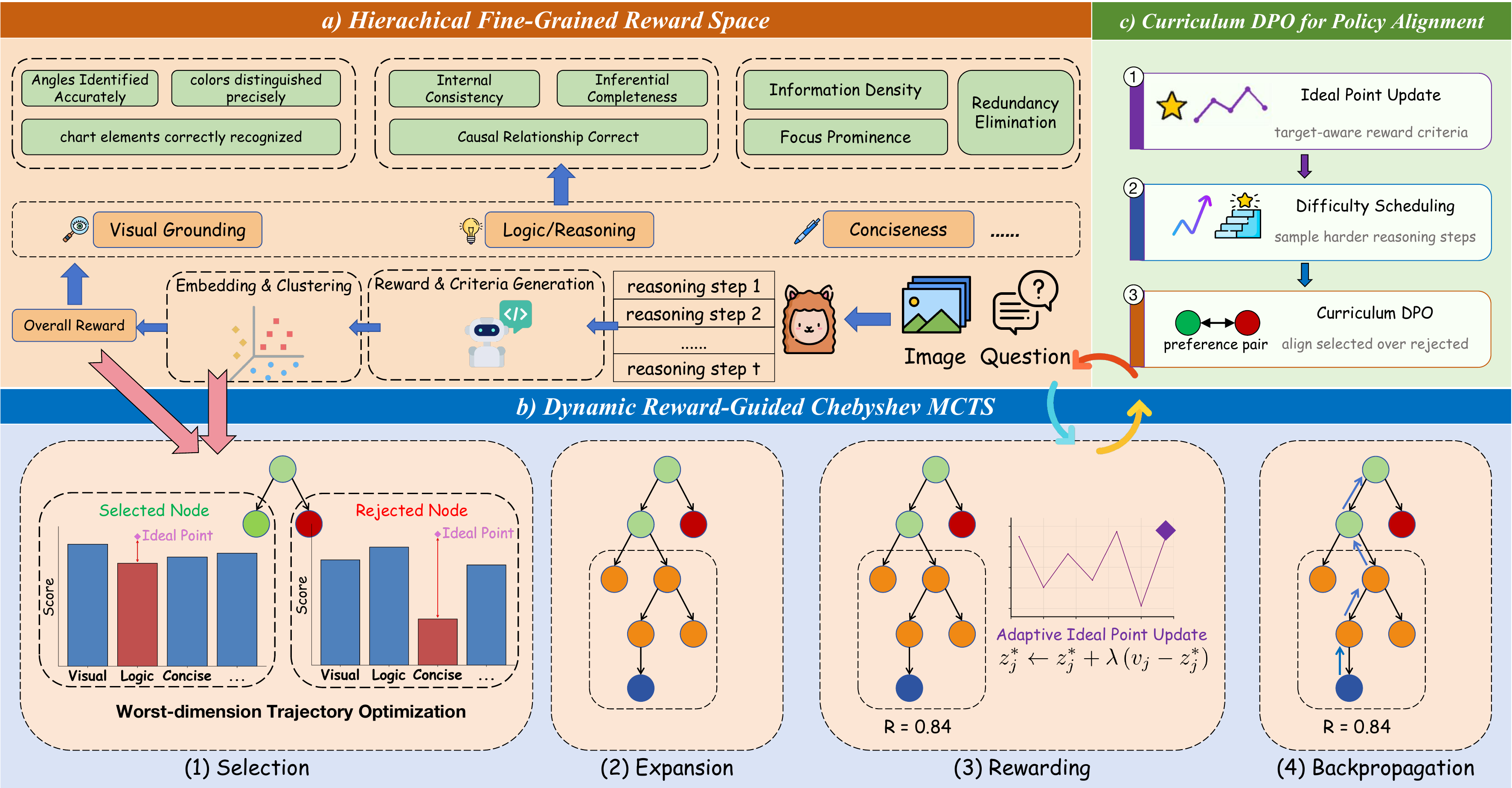}
\caption{Overview of MMS-PRM.
The framework consists of three components: (a) a hierarchical, fine-grained reward space that decomposes multimodal reasoning quality into interpretable dimensions and assigns step-level rewards; (b) a Chebyshev-guided Monte Carlo Tree Search (MCTS) that explores reasoning
trajectories by explicitly prioritizing the worst-performing dimension; and (c) a curriculum-style Direct Preference Optimization (DPO) that transfers balanced reasoning behaviors discovered by search into the policy. Together, these components form a closed-loop process that enforces
non-compensatory, balanced multimodal reasoning.}
\label{fig:method}
\end{figure*}

Multimodal reasoning differs from pure-text reasoning in that the model must keep every intermediate step visually grounded, logically coherent, and eventually answer-correct. This requires supervising both single-step quality and the dynamics from steps to the full trajectory. Simple final-answer supervision may overlook intermediate reasoning errors, while only step-level supervision may produce locally sound steps that fail to compose into a globally valid reasoning chain. To cope with this, we build a closed-loop alignment framework consisting of three parts: (1) a hierarchical, fine-grained reward space that models multimodal reasoning quality at the single-step level via multi-dimensional rewards; (2) a reward-guided Chebyshev MCTS that dynamically searches for trajectories balancing all activated reward dimensions; and (3) a curriculum-style DPO that progressively aligns the MLLM policy from easy/short chains to hard/long chains. The whole pipeline is shown in Figure~\ref{fig:method}.\\

\subsection{Hierarchical Fine-Grained Reward Space}
\label{subsec:reward-space}
The first component turns raw model outputs into reusable, structured reward dimensions.

\paragraph{Candidate Criteria Generation.}
For each instance in the training set, we derive criteria based on the reasoning steps. Given an input \( x \) and a reasoning sequence \( \hat{y} \) produced by model sampling, we apply an automated analyzer \( J(\cdot) \), typically instantiated as a vision-language model, to extract at most 5 relevant criteria:
\begin{equation}
    J(x, \hat{y}) = \{ c_1, c_2, \dots \},
\end{equation}
These criteria assess the reasoning quality, addressing various factors such as visual alignment, semantic correctness, logical consistency, stepwise coherence, and conciseness.

\paragraph{Embedding and Clustering.}
The criteria gathered are embedded into a \( d \)-dimensional vector space:
\begin{equation}
    V(c_i) = [v^{(1)}_i, \dots, v^{(d)}_i],
\end{equation}
where d is the embedding dimensionality.
hierarchical clustering is performed to group similar criteria together, halting when the similarity between criteria drops below a certain threshold \( \xi \). This results in a hierarchical reward structure \( \mathcal{H} = \{ \mathcal{H}_1, \mathcal{H}_2, \dots \} \), where higher levels are broad (e.g., "visually grounded") and lower levels are more specific (e.g., "accurately references bar \#3").

\paragraph{Stepwise Reward Allocation.}
The dynamic process of reward assignment employs a reward tree \( T \) to assign stepwise reward signals. For each step \( \hat{y}(t) \) in the \( n \)-step reasoning trajectory 
\( \hat{y} = \{ \hat{y}(1), \hat{y}(2), \dots, \hat{y}(n) \} \), 
the corresponding reward is allocated dynamically.

Specifically, for the \( t \)-th step \( \hat{y}(t) \), a selection function first chooses a coarse-grained parent reward node 
\( r_{\text{parent}} \) from the top level of the reward tree \( T \), representing the primary risk dimension to be evaluated at this step.
Conditioned on \( \hat{y}(t) \) and \( r_{\text{parent}} \), an analysis function \( \Phi \) determines whether finer-grained evaluation is required, and if so, generates a set of candidate textual evaluation criteria:
\begin{equation}
    \mathcal{C}_t = \Phi(\hat{y}(t), r_{\text{parent}}), \quad 
\mathcal{C}_t = \{ c_t^1, c_t^2, \dots, c_t^{K_t} \}.
\end{equation}

Each criterion \( c_t^i \in \mathcal{C}_t \) is embedded into a \( d \)-dimensional semantic space using the embedding function \( V(\cdot) \).
To filter out criteria that are weakly related or semantically drifting away from the selected parent reward, we compute the cosine distance between each criterion embedding and the parent reward embedding:
\begin{equation}
    \delta_i^t = D\bigl( V(c_t^i), V(r_{\text{parent}}) \bigr),
\end{equation}
where \( D(\cdot) \) denotes cosine distance.

This distance-based filtering serves as a heuristic safeguard rather than a strict semantic entailment test, aiming to remove candidate criteria that are loosely related to the intended risk dimension or shift focus away from the core evaluation objective.
Accordingly, the set of activated reward nodes for step \( t \) is defined as:
\begin{equation}
    R_t = \{ r_t^i \mid c_t^i \in \mathcal{C}_t,\; \delta_i^t \leq \zeta \},
\end{equation}
where each \( r_t^i \) denotes the reward tree node associated with criterion \( c_t^i \), and \( \zeta \) is a predefined distance threshold.

Finally, each activated reward node \( r_t^i \in R_t \) is scored using a step-level scoring function \( S(\cdot) \):
\begin{equation}
    s_i(t) = S\bigl( \hat{y}(t), r_t^i \bigr),
\end{equation}

All step-level reward scores produced by the scoring function \( S(\cdot) \) are initially defined on a discrete scale of \( [1, 10] \).
To ensure consistency across heterogeneous reward dimensions, we linearly normalize each reward dimension into \( [0, 1] \) before any trajectory aggregation or scalarization is applied.
Specifically, a raw score \( s \in [1, 10] \) is mapped to \( \tilde{s} = (s - 1) / 9 \).

\subsection{Dynamic Reward-Guided Chebyshev MCTS}
\label{subsec:drc-mcts}
The second component uses search to explore trajectories that are simultaneously good across all activated dimensions, where the reward space assigns a multi-dimensional reward vector to each reasoning step rather than a single scalar or final outcome.

\paragraph{Chebyshev scalarization.}
Given a reward vector $v = (v_1, \dots, v_M)$ and an adaptive ideal point  $z^\ast = (z^\ast_1, \dots, z^\ast_M)$, we first consider the standard Chebyshev scalarization,
defined as:
\begin{equation}
    g_{\text{cheby}}(v; z^\ast) = - \max_j \bigl( z^\ast_j - v_j \bigr).
    \label{eq:cheby}
\end{equation}
However, multimodal reasoning tasks often involve noisy reward signals, where a single isolated misclassification by the reward model could excessively penalize a valid trajectory under the strict min-max criterion. To mitigate this noise and enhance robustness, we employ the \emph{Augmented Chebyshev Scalarization}:
\begin{equation}
    g_{\text{aug}}(v; z^{*}) = g_{\text{cheby}}(v; z^\ast) - \rho \sum_{j=1}^{M}(z_{j}^{*} - v_{j})
    \label{eq:aug_chebyshev}
\end{equation}
where $\rho$ is a small positive constant serving as a regularization term. This formulation preserves 1. focus on the worst dimension due to the chebyshev term; 2. for trajectories that are good on all aspects, the augmented term can capture their subtle differences and identify the best candidate

\paragraph{MCTS Implementation.}
The MCTS algorithm constructs a tree of reasoning steps, where each node corresponds to a partial reasoning trajectory. In a nutshell, the algorithm dynamically searches for the best trajectory based on Chebyshev scalarized rewards:

\begin{itemize}
    \item \textbf{Selection:} Starting from the root, the algorithm traverses the tree by selecting the child node that maximizes the Upper Confidence Bound applied to Trees (UCT) score:
    \begin{equation}
        \text{UCT}(n) = g_{\text{aug}}(R_{\text{step}}(n); z^\ast) + c \sqrt{ \frac{\log N(\text{parent}(n))}{N(n)} },
    \end{equation}
    where $R_{\text{step}}(n)$ denote the $M$-dimensional reward vector associated with node $n$, while the scalarized step quality is always obtained via $g_{\text{aug}}(\cdot)$. $c$ controls the exploration-exploitation trade-off. 
    Here, N(n) represents the number of node n has been visited, while $N(\text{parent}(n))$ denotes the visit count of its parent node. The next node is selected via $n_{\text{next}} = \mathop{\arg\max}_{n' \in \text{children}(n)} \text{UCT}(n')$.
    
    \item \textbf{Expansion:} Upon reaching a leaf node, we expand the tree by sampling $k$ candidate steps $\{y^{(t)}_1, \dots, y^{(t)}_k\}$ from the current policy $\pi_{\theta}(y^{(t)} | x, y^{(<t)})$. Each candidate instantiates a new child node.
    
    \item \textbf{Evaluation:} For a newly expanded node $n$, we evaluate its immediate quality. Instead of random rollouts, we compute the fine-grained process reward vector $R_{\text{step}}(n)$. We then apply the Augmented Chebyshev Scalarization $g_{\text{aug}}(\cdot)$ to condense this vector into a scalar, strictly penalizing the worst-performing dimension while retaining discrimination via regularization.
    
    \item \textbf{Backpropagation:} The scalarized value and visit counts are propagated from the leaf to the root. For every node along the path, $N(n)$ is incremented and statistics are updated, dynamically guiding the search toward paths that are balanced across all reward dimensions.
\end{itemize}

\paragraph{Adaptive Ideal Point Update.}
In our framework, the ideal point is updated according to the
worst-performing reward dimension observed in the current MCTS
simulation. Given the current ideal point
$z^{*}=(z^{*}_{1},\ldots,z^{*}_{M})$ and the reward vector
$v=(v_{1},\ldots,v_{M})$, we compute the dimension-wise gap between
the ideal point and the observed reward:
\begin{equation}
\Delta_j = z^{*}_{j} - v_j, \quad j=1,\ldots,M .
\end{equation}

The worst-performing dimension is selected as the dimension with the
largest gap:
\begin{equation}
j^{*} = \arg\max_{j \in \{1,\ldots,M\}} \Delta_j .
\end{equation}

The maximum gap is then used as the adaptive signal for the next-round
ideal point:
\begin{equation}
\Delta_{\max}
=
\Delta_{j^{*}}
=
\max_{j \in \{1,\ldots,M\}}
\left(z^{*}_{j}-v_j\right).
\end{equation}

Finally, we broadcast this worst-dimension gap to construct the next
ideal-point vector:
\begin{equation}
z^{*}_{\mathrm{next}}
=
\Delta_{\max}\mathbf{1}_{M}.
\end{equation}

This update directly uses the largest deviation from the ideal point as
the next reference signal, thereby forcing the following MCTS iteration
to concentrate on the currently weakest dimension. Unlike
coordinate-wise exponential moving average updates, the proposed rule
does not allow strong dimensions to dilute the failure of weak ones.
Therefore, it better matches the non-compensatory objective of
Chebyshev-guided search and strengthens the worst-dimension
optimization principle of MMS-PRM.

\subsection{Curriculum DPO for Policy Alignment}
\label{subsec:curr-dpo}
After applying the MCTS algorithm, we obtain an ideal point \(z^*\) together with a set of candidate trajectories for each input sample.
The third component then converts these searched trajectories into preference data and aligns the policy using an easy-to-hard curriculum.

\paragraph{Score fusion and pair construction.}
Our method allows us to evaluate the overall quality of any reasoning trajectory \(Y\) via a fused scalar score \(G(Y)\).
Based on this evaluation, preference pairs \((Y^+, Y^-)\) are constructed from the searched trajectories produced by MCTS, satisfying $G(Y^+) - G(Y^-) \ge \delta$.
Giving a reasoning trajectory $Y = \{y^{(1)}, y^{(2)}, \dots, y^{(n)}\}$, we define the trajectory-level reward vector $\mathbf{R}_{\text{traj}}(Y) \in \mathbb{R}^M$ as:
\begin{equation}
\mathbf{R}_{\text{traj}}(Y) = \frac{1}{n} \sum_{t=1}^{n} \mathbf{R}_{\text{step}}(y^{(t)})
\end{equation}
where each $\mathbf{R}_{\text{step}}(y^{(t)})$ represents the $M$-dimensional reward scores (e.g., logical consistency, visual alignment) at step $t$. This definition ensures that the trajectory reward maintains the same dimensionality and physical meaning as the step-wise rewards.
After $K$ simulations, each trajectory $Y$ has a fused score
\begin{equation}
    \bar{G}(Y) = \eta \, g_{\text{aug}}(R_{\text{traj}}(Y)) + (1-\eta) R_{\text{ans}}(Y),
\end{equation}
where $\eta \in [0,1]$ is a balancing coefficient. $R_{ans}(n)$ denotes the outcome correctness reward, defined as $1$ if the reasoning chain at terminal node $n$ matches the ground truth answer, and $0$ otherwise. For non-terminal nodes, $R_{ans}(n)$ is set to $0$.

Finally, given multiple reasoning trajectories sampled for the same input $x$, we induce preference pairs $(Y^+, Y^-)$ for DPO-style optimization, where the trajectory with a higher aggregated score is treated as preferred, i.e.,
\begin{equation}
\bar{G}(Y^+) - \bar{G}(Y^-) \ge \delta.
\end{equation}

\paragraph{Warm-up and DPO.}
Given the constructed preference pairs $(Y^+, Y^-)$ derived from the search process, 
this stage aims to align the policy with step-level reasoning preferences through a two-phase training procedure.
Specifically, we first warm up the policy using supervised fine-tuning (SFT) to stabilize generation,
and then apply Direct Preference Optimization (DPO) to directly optimize the policy toward preferred trajectories.

\begin{equation}
    \mathcal{L}_{\text{SFT}}(\pi_\theta) = - \mathbb{E}_{(x,Y)} \sum_{t=1}^T \log \pi_\theta(y^{(t)} \mid x, y^{(<t)}).
\end{equation}
Then we apply DPO with a reference model $\pi_{\text{ref}}$:
\begin{equation}
L(\theta) = -\mathbb{E}_{x, \hat{y}^{<t}, (\hat{y}^{+}, \hat{y}^{-}) \in V} \left[ \log \sigma(F^+ - F^-) \right]
\end{equation}
where,
\begin{equation}
F^+ = \beta \log \frac{\pi_{\theta}(\hat{y}^{+} | x; \hat{y}^{<t})}{\pi_{\text{ref}}(\hat{y}^{+} | x; \hat{y}^{<t})}
\end{equation}

\begin{equation}
F^- = \beta \log \frac{\pi_{\theta}(\hat{y}^{-} | x; \hat{y}^{<t})}{\pi_{\text{ref}}(\hat{y}^{-} | x; \hat{y}^{<t})}
\end{equation}

\paragraph{Difficulty scheduling.}
We define the difficulty of a reasoning trajectory based on two intuitive factors: 
its overall reward quality and its reasoning length.

A trajectory whose reward vector is closer to the ideal point $z^*$ corresponds to higher overall reward,
and is therefore considered easier.
Similarly, shorter reasoning trajectories generally indicate simpler reasoning processes and are also treated as easier cases.

Taking both factors into account, we define the difficulty score of a trajectory $Y$ as a linear combination of its normalized depth and its distance to the ideal point:
\begin{equation}
D(Y) = \alpha \frac{\text{depth}(Y)}{T_{\max}} 
+ (1-\alpha) \frac{\| z^* - \mathbf{R}_{\text{traj}}(Y) \|_2}{\sqrt{M}},
\end{equation}
where $T_{\max}$ denotes the maximum allowed number of reasoning steps, 
$M$ is the number of reward dimensions, and $\mathbf{R}_{\text{traj}}(Y)$ represents the aggregated reward vector of trajectory $Y$.
The $L_2$ distance measures how far the trajectory performance deviates from the ideal point, and dividing by $\sqrt{M}$ normalizes the term to $[0,1]$.

By construction, a smaller value of $D(Y)$ indicates an easier trajectory, while larger values correspond to harder cases.
We use this difficulty score to perform curriculum-style difficulty scheduling during training, 
prioritizing easier trajectories in early stages and gradually incorporating harder ones as training progresses.

\paragraph{Closed-loop training.}
Searching with the current policy yields better-balanced trajectories; these trajectories give us preference data with graded difficulty; the updated policy, in turn, improves the quality of the search. This closes the loop and steadily pushes the model toward high-quality multimodal reasoning.

\section{Experiment}
\begin{table*}[ht]
    \centering
    \small
    \begin{tabular}{lcccccccc}
        \toprule
        \textbf{Method} & \textbf{Size} & \textbf{MathVista} & \textbf{MMStar} & \textbf{MMMU} & \textbf{M3CoT} & \textbf{AI2D} & \textbf{ChartQA} & \textbf{Average} \\
        \midrule
        LLaVA-1.5 \cite{liu2024improved} & 13B & 27.6 & 32.8 & - & 39.5 & - & - & - \\
        IXC-2.5 \cite{zhang2024internlm} & 7B & 63.7 & 59.9 & 42.9 & - & 81.5 & 82.2 & - \\
        Ovis1.5-LLaMA3 \cite{lu2024ovis} & 8B & 63.0 & 57.3 & 48.3 & - & - & 76.4 & - \\
        LLaVA-CoT \cite{xu2025llava} & 11B & 54.8 & 57.6 & - & - & 78.7 & - & - \\
        LlamaV-o1 \cite{thawakar2025llamav} & 11B & 54.4 & 59.5 & - & - & 81.2 & - & - \\
        Vision-R1-LlamaV \cite{huang2025vision} & 11B & 62.7 & 61.4 & - & - & - & 83.9 & - \\
        R1-VL \cite{zhang2025r1} & 7B & 63.5 & 60.0 & - & - & - & 83.9 & - \\
        R1-Onevision \cite{yang2025r1} & 7B & 64.1 & - & - & - & - & - & - \\
        MiniCPM-V-2-6 \cite{yao2024minicpm} & 8B & 60.6 & 57.5 & 49.8 & 56.0 & 82.1 & 79.4 & 64.2 \\
        LLaVA-OneVision \cite{li2024llava} & 7B & 63.2 & 61.7 & 48.8 & 52.3 & 81.4 & 80.0 & 64.6 \\
        Qwen2-VL \cite{wang2024qwen2} & 7B & 58.2 & 60.7 & 53.7 & 57.8 & 83.0 & 77.4 & 65.1 \\
        Insight-V \cite{dong2025insight} & 7B & 59.9 & 61.5 & 50.2 & 61.5 & 79.8 & 81.5 & 65.7 \\
        InternVL-2.5 \cite{chen2024expanding} & 8B & 62.2 & 59.6 & 54.1 & 62.4 & 84.5 & 84.9 & 68.0 \\
        \midrule
        LLaVA-NeXT-LLaMA3 \cite{liu2024llavanext} & 8B & 45.9 & 43.1 & 36.9 & 45.6 & 71.5 & 69.4 & 52.1 \\
        LLaVA-NeXT-LLaMA3 + SFT & 8B & 51.4 & 54.7 & 39.6 & 67.4 & 76.1 & 75.7 & 60.8 \\
        InternVL2.5-MPO \cite{wang2024enhancing} & 8B & 65.0 & 60.7 & 53.8 & 67.5 & 84.2 & 85.0 & 69.4 \\
        InternVL2.5-MPO + SFT & 8B & 65.9 & 61.0 & 53.7 & 75.7 & 81.6 & \textbf{88.3} & 71.0 \\
        \midrule
        MMS-PRM & 8B & \textbf{67.5} & \textbf{65.2} & \textbf{54.2} & \textbf{79.7} & \textbf{84.2} & 87.2 & \textbf{73.0} \\
        \bottomrule
    \end{tabular}
    \caption{Comparison with open-source VLMs. We evaluate our method on six benchmarks covering both general and task-specific reasoning capacities. Our method has consistently strong performances across these benchmarks, surpassing different baselines and other state-of-the-art VLMs by large margins. The items in bold and underlined respectively represent the first or second highest scores.}
\end{table*}

\subsection{Experimental Setup}

\subsubsection{Implementation details}
We employ InternVL-2.5-MPO\cite{chen2024expanding} as our base Vision-Language Model (VLM). The supervised fine-tuning is conducted on ShareGPT-step-300K for one epoch. Qwen2.5-VL-32B-Instruct is adopted as the backbone of the Reward and Criteria Generation Model.
The BAAI/bge-en-icl model~\cite{lee2025nv} is used to construct the embedding space $V \in \mathbb{R}^d$, where each element corresponds to a vector representation of a generated criterion, and the dimensionality is set to $d=4096$.
To organize these criteria at multiple granularities, hierarchical clustering is performed using the BIRCH algorithm~\cite{zhang1996birch},  resulting in a hierarchical structure $H$ over the criterion embeddings.
 For the training time, SFT takes about 9 hours on a single A800 node, and three rounds of Curriculum DPO takes about 18 hours in total on a 4 * A100 node. We set branch factor $k=3$ for a good balance between exploration and computation time. We set Search Depth $D=10$ to limit the search space while maintaining adequate coverage of reasoning paths. We empirically set Reward Parameters $\eta=0.5$, $\rho=0.1$, and $\lambda=0.2$ to optimize reasoning quality while avoiding overfitting to any single dimension.

\subsubsection{Benchmarks}
We evaluate our method on several representative benchmarks, spanning a wide range of tasks that demand complex multimodal reasoning abilities. For most of these benchmarks, performance is measured by accuracy (or relaxed accuracy)—i.e., the ratio of correct predictions to total examples—unless specified otherwise.
MathVista~\cite{lu2024mathvista} is a well-known benchmark for multimodal mathematical reasoning, focusing on tasks such as plane geometry, functions, and visual puzzles.
MMMU~\cite{yue2024mmmu} evaluates multimodal reasoning across diverse university-level disciplines.
ChartQA~\cite{masry2022chartqa} aims to assess logical and numerical reasoning over chart-based visual data.
MMStar~\cite{chen2024we} focuses on diverse multimodal problems that require fine-grained visual understanding.
M3CoT~\cite{shao2024visual} evaluates multimodal chain-of-thought reasoning capabilities.
AI2D~\cite{kembhavi2016diagram} is designed to test scientific diagram understanding and reasoning.

\subsection{Main Results}

Table~1 reports the comparison between MMS-PRM and state-of-the-art open-source VLMs across six multimodal reasoning benchmarks. Our method consistently achieves performance gains across the majority of benchmarks
Starting from InternVL2.5-MPO, supervised fine-tuning yields limited gains, indicating that performance is close to saturation under standard training. In contrast, introducing MMS-PRM leads to consistent improvements across all benchmarks, demonstrating the effectiveness of search-enhanced, multi-dimensional process supervision beyond conventional SFT.
The gains are particularly pronounced on reasoning-intensive benchmarks such as M3CoT and MathVista, which require long-horizon, multi-step reasoning. This suggests that MMS-PRM is especially effective at refining complex multimodal reasoning behaviors rather than optimizing for final-answer correctness alone.

Overall, these results show that explicitly balancing multiple reward dimensions at the process level remains a powerful mechanism for improving strong vision-language models.

\subsection{Efficiency and Generalization Analysis}
To further validate the practicality of MMS-PRM, we compare it against state-of-the-art multimodal PRMs that require extensive supervised fine-tuning. As shown in Table~\ref{tab:prm_efficiency}, while training-based methods such as Visual-PRM and SVIP achieve slightly superior accuracy by optimizing billions of parameters on large-scale step-level datasets, MMS-PRM remains highly competitive despite utilizing a training-free reward mechanism. This design avoids the need for training a separate reward model, distinguishing it from baselines that require optimizing billions of parameters for step-level supervision. The minor performance trade-off is significantly offset by our model’s superior deployment efficiency and its robust generalization across diverse benchmarks, as it avoids the risk of overfitting to specific training distributions or reward hacking. Consequently, MMS-PRM offers a more resource-efficient and scalable paradigm for real-world multimodal reasoning.
\begin{table}[ht]
\centering
\small
\setlength{\tabcolsep}{3pt} 
\begin{tabular*}{\columnwidth}{@{\extracolsep{\fill}}lcccc}
\toprule
\textbf{Method} & \textbf{PRM Trainable Params} & \textbf{MathVista} & \textbf{MMMU} \\
\midrule
Visual-PRM & 7B & 68.5 & 60.2 \\
SVIP & 7B & 61.6 & 61.3 \\
URSA & 8B & 64.9 & - \\
VL-PRM300K & 8B & 65.7 & 56.5 \\
\midrule
\textbf{MMS-PRM} & \textbf{0M} & 67.5 & 54.2 \\
\bottomrule
\end{tabular*}
\caption{Comparison with state-of-the-art multimodal PRMs. Our method achieves competitive performance using a training-free process reward model, whereas baselines require extensive supervision to train their reward models.}
\label{tab:prm_efficiency}
\end{table}

\subsection{Ablation Study}

We perform ablation studies on the M3CoT validation set to analyze the contribution of each component in MMS-PRM. As shown in Table~\ref{tab:ablation}, introducing the hierarchical fine-grained process rewards consistently improves the SFT baseline, indicating the effectiveness of step-level supervision for multimodal reasoning.
Further incorporating reward-guided Chebyshev MCTS leads to additional gains, demonstrating that explicitly searching for balanced reasoning trajectories is more effective than direct optimization. Compared with weighted-sum MCTS, Chebyshev scalarization achieves better performance, suggesting that penalizing the weakest reward dimension is crucial for multimodal reasoning.
Finally, the full MMS-PRM framework, which combines hierarchical rewards, Chebyshev MCTS, and curriculum-style DPO, achieves the best performance. In contrast, removing MCTS and applying DPO alone results in a noticeable drop, highlighting the necessity of search-based trajectory optimization.

\begin{table}[t]
\centering
\footnotesize
\setlength{\tabcolsep}{4pt}
\renewcommand{\arraystretch}{1.1}
\begin{tabular}{lp{2.2cm}}
\toprule
\textbf{Configuration} & \textbf{Acc. (\%)} \\
\midrule
Baseline (SFT) & 67.4 \\
+ Hierarchical reward & 70.1 \\
+ Reward + Chebyshev MCTS & 73.6 \\
MMS-PRM (full) & \textbf{79.7} \\
\midrule
DPO only (w/o MCTS) & 71.2 \\
Weighted-sum MCTS & 75.3 \\
\bottomrule
\end{tabular}
\caption{Ablation study results on the M3CoT validation set.}
\label{tab:ablation}
\end{table}

\subsection{Impact of Reward Count}
We further investigate the impact of supervision density by analyzing the average reward counts. In this experiment, we control the density of process supervision by explicitly constraining the number of output rewards in the prompt provided to the criteria generation model. Specifically, we adjust the instructions to request a varying number of criteria—ranging from a strict constraint (e.g., "output only the single most critical criterion") to a relaxed setting that encourages generating comprehensive criteria across multiple dimensions. As shown in Table~\ref{tab:reward-count-ablation}, the reasoning performance consistently improves as the average number of activated process rewards increases from 1.0 to 4.22.
Here, the average values are induced by prompt-level constraints on the number of generated criteria (e.g., exactly 1, 1–3, $\leq$5, and 3–5), rather than being directly set. This trend indicates that MMS-PRM benefits significantly from joint multi-dimensional constraints.
\begin{table}[ht]
\centering
\small
\setlength{\tabcolsep}{5pt}
\begin{tabular}{ccccc}
\toprule
\textbf{Avg Rewards / Step} 
& \textbf{MathVista} 
& \textbf{M3CoT} 
& \textbf{ChartQA} 
& \textbf{Avg.} \\
\midrule
1.0 
& 65.8 
& 74.1 
& 86.4 
& 74.8 \\

2.1 
& 66.9 
& 76.3 
& 87.1 
& 76.1 \\

3.4 
& 68.2 
& 78.8 
& \textbf{88.4} 
& 77.8 \\

4.2  
& \textbf{70.1} 
& \textbf{79.7} 
& 87.2 
& \textbf{79.6} \\
\bottomrule
\end{tabular}
\caption{Effect of reward dimensionality measured by the average number of activated rewards per reasoning step.}
\label{tab:reward-count-ablation}
\end{table}
\subsection{Worst-Dimension Analysis}
We analyze the minimum reward dimension $v_{\min}=\min_j v_j$ of final trajectories generated by Chebyshev-guided MCTS and weighted-sum MCTS under identical rollout budgets and reward models. As shown in Figure~3, Chebyshev scalarization consistently shifts the distribution of $v_{\min}$ toward higher values, indicating that the weakest reward dimension is significantly improved.
In contrast, weighted-sum MCTS exhibits a heavier tail in the low-$v_{\min}$ region, suggesting that severe failures in individual dimensions can be masked by strong performance in others. This confirms that Chebyshev-based optimization effectively enforces balanced multimodal reasoning by penalizing worst-dimension collapse rather than allowing compensation across dimensions.
\begin{figure}[ht!]
\centering
\includegraphics[width=\linewidth]{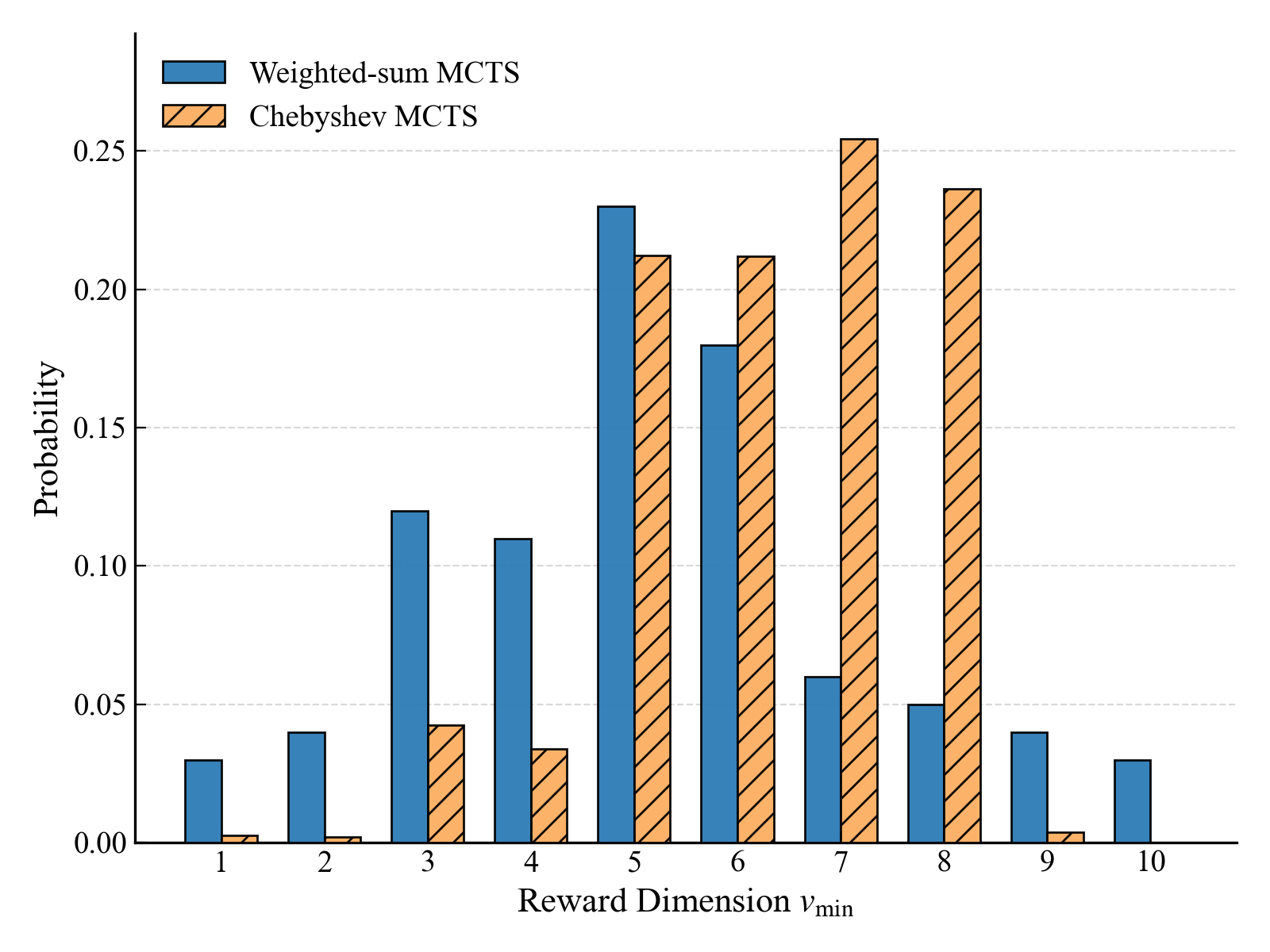}
\caption{
Distribution of the minimum reward dimension $v_{\min}$ across reasoning trajectories
}
\end{figure}

\section{Conclusion}
This paper introduced MMS-PRM, which replaces superficial scalar rewards with fine-grained, multi-dimensional supervision via Chebyshev-scalarized MCTS. Our framework effectively balances visual and logical reasoning, achieving competitive results and robust generalization. These findings highlight the critical role of structured process rewards and search mechanisms in advancing reliable, interpretable multimodal reasoning. 

\section*{Acknowledgments}

This work was supported by the National Key R\&D Program of China
under Grant No.~2024YFC3308101, for the project ``Long- and Short-Term
Holographic Profiling of Bond Investors Based on Trading Behavior Features'',
led by Prof.~Huaping Zhang.

\bibliographystyle{named}
\bibliography{ijcai26}

@String{Computer = "{IEEE} Computer" }

@String{Springer = "Springer-Verlag" }

@inproceedings{qi2025mutual,
  title={Mutual reasoning makes smaller LLMs stronger problem-solver},
  author={Qi, Zhenting and Ma, Mingyuan and Xu, Jiahang and Zhang, Li Lyna and Yang, Fan and Yang, Mao},
  booktitle={International Conference on Learning Representations},
  volume={2025},
  pages={20788--20807},
  year={2025}
}

@article{wu2024beyond,
  title={Beyond examples: High-level automated reasoning paradigm in in-context learning via mcts},
  author={Wu, Jinyang and Feng, Mingkuan and Zhang, Shuai and Che, Feihu and Wen, Zengqi and Liao, Chonghua and Tao, Jianhua},
  journal={arXiv preprint arXiv:2411.18478},
  year={2024}
}

@inproceedings{dong2025progressive,
  title={Progressive multimodal reasoning via active retrieval},
  author={Dong, Guanting and Zhang, Chenghao and Deng, Mengjie and Zhu, Yutao and Dou, Zhicheng and Wen, Ji-Rong},
  booktitle={Proceedings of the 63rd Annual Meeting of the Association for Computational Linguistics (Volume 1: Long Papers)},
  pages={3579--3602},
  year={2025}
}

@article{yao2026mulberry,
  title={Mulberry: Empowering mllm with o1-like reasoning and reflection via collective monte carlo tree search},
  author={Yao, Huanjin and Huang, Jiaxing and Wu, Wenhao and Zhang, Jingyi and Wang, Yibo and Liu, Shunyu and Wang, Yingjie and Song, Yuxin and Feng, Haocheng and Shen, Li and others},
  journal={Advances in Neural Information Processing Systems},
  volume={38},
  pages={29918--29952},
  year={2026}
}

@inproceedings{yue2024mmmu,
  title={Mmmu: A massive multi-discipline multimodal understanding and reasoning benchmark for expert agi},
  author={Yue, Xiang and Ni, Yuansheng and Zhang, Kai and Zheng, Tianyu and Liu, Ruoqi and Zhang, Ge and Stevens, Samuel and Jiang, Dongfu and Ren, Weiming and Sun, Yuxuan and others},
  booktitle={Proceedings of the IEEE/CVF conference on computer vision and pattern recognition},
  pages={9556--9567},
  year={2024}
}

@inproceedings{dong2025insight,
  title={Insight-v: Exploring long-chain visual reasoning with multimodal large language models},
  author={Dong, Yuhao and Liu, Zuyan and Sun, Hai-Long and Yang, Jingkang and Hu, Winston and Rao, Yongming and Liu, Ziwei},
  booktitle={Proceedings of the Computer Vision and Pattern Recognition Conference},
  pages={9062--9072},
  year={2025}
}

@inproceedings{zhang2025improve,
  title={Improve vision language model chain-of-thought reasoning},
  author={Zhang, Ruohong and Zhang, Bowen and Li, Yanghao and Zhang, Haotian and Sun, Zhiqing and Gan, Zhe and Yang, Yinfei and Pang, Ruoming and Yang, Yiming},
  booktitle={Proceedings of the 63rd Annual Meeting of the Association for Computational Linguistics (Volume 1: Long Papers)},
  pages={1631--1662},
  year={2025}
}

@article{li2024llava,
  title={Llava-onevision: Easy visual task transfer},
  author={Li, Bo and Zhang, Yuanhan and Guo, Dong and Zhang, Renrui and Li, Feng and Zhang, Hao and Zhang, Kaichen and Zhang, Peiyuan and Li, Yanwei and Liu, Ziwei and others},
  journal={arXiv preprint arXiv:2408.03326},
  year={2024}
}

@article{snell2024scaling,
  title={Scaling llm test-time compute optimally can be more effective than scaling model parameters},
  author={Snell, Charlie and Lee, Jaehoon and Xu, Kelvin and Kumar, Aviral},
  journal={arXiv preprint arXiv:2408.03314},
  year={2024}
}

@article{luo2024improve,
  title={Improve mathematical reasoning in language models by automated process supervision},
  author={Luo, Liangchen and Liu, Yinxiao and Liu, Rosanne and Phatale, Samrat and Guo, Meiqi and Lara, Harsh and Li, Yunxuan and Shu, Lei and Zhu, Yun and Meng, Lei and others},
  journal={arXiv preprint arXiv:2406.06592},
  year={2024}
}

@article{guan2025rstar,
  title={RStar-math: Small LLMs can master math reasoning with self-evolved deep thinking},
  author={Guan, Xinyu and Zhang, Li Lyna and Liu, Yifei and Shang, Ning and Sun, Youran and Zhu, Yi and Yang, Fan and Yang, Mao},
  journal={arXiv preprint arXiv:2501.04519},
  year={2025}
}

@inproceedings{wang2024math,
  title={Math-shepherd: Verify and reinforce llms step-by-step without human annotations},
  author={Wang, Peiyi and Li, Lei and Shao, Zhihong and Xu, Runxin and Dai, Damai and Li, Yifei and Chen, Deli and Wu, Yu and Sui, Zhifang},
  booktitle={Proceedings of the 62nd Annual Meeting of the Association for Computational Linguistics (Volume 1: Long Papers)},
  pages={9426--9439},
  year={2024}
}

@inproceedings{wang2024multi,
  title={Multi-step problem solving through a verifier: An empirical analysis on model-induced process supervision},
  author={Wang, Zihan and Li, Yunxuan and Wu, Yuexin and Luo, Liangchen and Hou, Le and Yu, Hongkun and Shang, Jingbo},
  booktitle={Findings of the Association for Computational Linguistics: EMNLP 2024},
  pages={7309--7319},
  year={2024}
}

@inproceedings{liu2024improved,
  title={Improved baselines with visual instruction tuning},
  author={Liu, Haotian and Li, Chunyuan and Li, Yuheng and Lee, Yong Jae},
  booktitle={Proceedings of the IEEE/CVF conference on computer vision and pattern recognition},
  pages={26296--26306},
  year={2024}
}

@article{zhang2024internlm,
  title={Internlm-xcomposer-2.5: A versatile large vision language model supporting long-contextual input and output},
  author={Zhang, Pan and Dong, Xiaoyi and Zang, Yuhang and Cao, Yuhang and Qian, Rui and Chen, Lin and Guo, Qipeng and Duan, Haodong and Wang, Bin and Ouyang, Linke and others},
  journal={arXiv preprint arXiv:2407.03320},
  year={2024}
}

@article{lu2024ovis,
  title={Ovis: Structural embedding alignment for multimodal large language model},
  author={Lu, Shiyin and Li, Yang and Chen, Qing-Guo and Xu, Zhao and Luo, Weihua and Zhang, Kaifu and Ye, Han-Jia},
  journal={arXiv preprint arXiv:2405.20797},
  year={2024}
}

@inproceedings{xu2025llava,
  title={Llava-cot: Let vision language models reason step-by-step},
  author={Xu, Guowei and Jin, Peng and Wu, Ziang and Li, Hao and Song, Yibing and Sun, Lichao and Yuan, Li},
  booktitle={Proceedings of the IEEE/CVF International Conference on Computer Vision},
  pages={2087--2098},
  year={2025}
}

@inproceedings{thawakar2025llamav,
  title={Llamav-o1: Rethinking step-by-step visual reasoning in llms},
  author={Thawakar, Omkar and Dissanayake, Dinura and More, Ketan Pravin and Thawkar, Ritesh and Heakl, Ahmed and Ahsan, Noor and Li, Yuhao and Zumri, Ilmuz Zaman Mohammed and Lahoud, Jean and Anwer, Rao Muhammad and others},
  booktitle={Findings of the Association for Computational Linguistics: ACL 2025},
  pages={24290--24315},
  year={2025}
}

@article{huang2025vision,
  title={Vision-r1: Incentivizing reasoning capability in multimodal large language models},
  author={Huang, Wenxuan and Jia, Bohan and Zhai, Zijie and Cao, Shaosheng and Ye, Zheyu and Zhao, Fei and Xu, Zhe and Tang, Xu and Hu, Yao and Lin, Shaohui},
  journal={arXiv preprint arXiv:2503.06749},
  year={2025}
}

@article{zhang2025r1,
  title={R1-vl: Learning to reason with multimodal large language models via step-wise group relative policy optimization},
  author={Zhang, Jingyi and Huang, Jiaxing and Yao, Huanjin and Liu, Shunyu and Zhang, Xikun and Lu, Shijian and Tao, Dacheng},
  journal={arXiv preprint arXiv:2503.12937},
  year={2025}
}

@inproceedings{yang2025r1,
  title={R1-onevision: Advancing generalized multimodal reasoning through cross-modal formalization},
  author={Yang, Yi and He, Xiaoxuan and Pan, Hongkun and Jiang, Xiyan and Deng, Yan and Yang, Xingtao and Lu, Haoyu and Yin, Dacheng and Rao, Fengyun and Zhu, Minfeng and others},
  booktitle={Proceedings of the IEEE/CVF International Conference on Computer Vision},
  pages={2376--2385},
  year={2025}
}

@article{yao2024minicpm,
  title={Minicpm-v: A gpt-4v level mllm on your phone},
  author={Yao, Yuan and Yu, Tianyu and Zhang, Ao and Wang, Chongyi and Cui, Junbo and Zhu, Hongji and Cai, Tianchi and Li, Haoyu and Zhao, Weilin and He, Zhihui and others},
  journal={arXiv preprint arXiv:2408.01800},
  year={2024}
}

@article{wang2024qwen2,
  title={Qwen2-vl: Enhancing vision-language model's perception of the world at any resolution},
  author={Wang, Peng and Bai, Shuai and Tan, Sinan and Wang, Shijie and Fan, Zhihao and Bai, Jinze and Chen, Keqin and Liu, Xuejing and Wang, Jialin and Ge, Wenbin and others},
  journal={arXiv preprint arXiv:2409.12191},
  year={2024}
}

@article{chen2024expanding,
  title={Expanding performance boundaries of open-source multimodal models with model, data, and test-time scaling},
  author={Chen, Zhe and Wang, Weiyun and Cao, Yue and Liu, Yangzhou and Gao, Zhangwei and Cui, Erfei and Zhu, Jinguo and Ye, Shenglong and Tian, Hao and Liu, Zhaoyang and others},
  journal={arXiv preprint arXiv:2412.05271},
  year={2024}
}

@misc{liu2024llavanext,
  title={Llavanext: Improved reasoning, ocr, and world knowledge},
  author={Liu, Haotian and Li, Chunyuan and Li, Yuheng and Li, Bo and Zhang, Yuanhan and Shen, Sheng and Lee, Yong Jae},
  year={2024}
}

@article{wang2024enhancing,
  title={Enhancing the reasoning ability of multimodal large language models via mixed preference optimization},
  author={Wang, Weiyun and Chen, Zhe and Wang, Wenhai and Cao, Yue and Liu, Yangzhou and Gao, Zhangwei and Zhu, Jinguo and Zhu, Xizhou and Lu, Lewei and Qiao, Yu and others},
  journal={arXiv preprint arXiv:2411.10442},
  year={2024}
}

@inproceedings{lee2025nv,
  title={Nv-embed: Improved techniques for training llms as generalist embedding models},
  author={Lee, Chankyu and Roy, Rajarshi and Xu, Mengyao and Raiman, Jonathan and Shoeybi, Mohammad and Catanzaro, Bryan and Ping, Wei},
  booktitle={International Conference on Learning Representations},
  volume={2025},
  pages={79310--79333},
  year={2025}
}

@article{zhang1996birch,
  title={BIRCH: an efficient data clustering method for very large databases},
  author={Zhang, Tian and Ramakrishnan, Raghu and Livny, Miron},
  journal={ACM sigmod record},
  volume={25},
  number={2},
  pages={103--114},
  year={1996},
  publisher={ACM New York, NY, USA}
}

@inproceedings{lu2024mathvista,
  title={Mathvista: Evaluating mathematical reasoning of foundation models in visual contexts},
  author={Lu, Pan and Bansal, Hritik and Xia, Tony and Liu, Jiacheng and Li, Chunyuan and Hajishirzi, Hannaneh and Cheng, Hao and Chang, Kai-Wei and Galley, Michel and Gao, Jianfeng},
  booktitle={International Conference on Learning Representations},
  volume={2024},
  pages={23439--23554},
  year={2024}
}

@inproceedings{masry2022chartqa,
  title={Chartqa: A benchmark for question answering about charts with visual and logical reasoning},
  author={Masry, Ahmed and Do, Xuan Long and Tan, Jia Qing and Joty, Shafiq and Hoque, Enamul},
  booktitle={Findings of the association for computational linguistics: ACL 2022},
  pages={2263--2279},
  year={2022}
}

@article{chen2024we,
  title={Are we on the right way for evaluating large vision-language models?},
  author={Chen, Lin and Li, Jinsong and Dong, Xiaoyi and Zhang, Pan and Zang, Yuhang and Chen, Zehui and Duan, Haodong and Wang, Jiaqi and Qiao, Yu and Lin, Dahua and others},
  journal={Advances in Neural Information Processing Systems},
  volume={37},
  pages={27056--27087},
  year={2024}
}

@inproceedings{kembhavi2016diagram,
  title={A diagram is worth a dozen images},
  author={Kembhavi, Aniruddha and Salvato, Mike and Kolve, Eric and Seo, Minjoon and Hajishirzi, Hannaneh and Farhadi, Ali},
  booktitle={European conference on computer vision},
  pages={235--251},
  year={2016},
  organization={Springer}
}

@article{shao2024visual,
  title={Visual cot: Advancing multi-modal language models with a comprehensive dataset and benchmark for chain-of-thought reasoning},
  author={Shao, Hao and Qian, Shengju and Xiao, Han and Song, Guanglu and Zong, Zhuofan and Wang, Letian and Liu, Yu and Li, Hongsheng},
  journal={Advances in Neural Information Processing Systems},
  volume={37},
  pages={8612--8642},
  year={2024}
}

@article{wang2025visualprm,
  title={Visualprm: An effective process reward model for multimodal reasoning},
  author={Wang, Weiyun and Gao, Zhangwei and Chen, Lianjie and Chen, Zhe and Zhu, Jinguo and Zhao, Xiangyu and Liu, Yangzhou and Cao, Yue and Ye, Shenglong and Zhu, Xizhou and others},
  journal={arXiv preprint arXiv:2503.10291},
  year={2025}
}

@article{luo2025ursa,
  title={Ursa: Understanding and verifying chain-of-thought reasoning in multimodal mathematics},
  author={Luo, Ruilin and Zheng, Zhuofan and Wang, Yifan and Yu, Yiyao and Ni, Xinzhe and Lin, Zicheng and Zeng, Jin and Yang, Yujiu},
  journal={arXiv e-prints},
  pages={arXiv--2501},
  year={2025}
}

@article{ong2025training,
  title={Training vision-language process reward models for test-time scaling in multimodal reasoning: Key insights and lessons learned},
  author={Ong, Brandon and Pala, Tej Deep and Toh, Vernon and Tjhi, William Chandra and Poria, Soujanya},
  journal={arXiv preprint arXiv:2509.23250},
  year={2025}
}

@inproceedings{gao2025benchmarking,
  title={Benchmarking multimodal cot reward model stepwise by visual program},
  author={Gao, Minghe and Liu, Xuqi and Yue, Zhongqi and Wu, Yang and Chen, Shuang and Li, Juncheng and Tang, Siliang and Wu, Fei and Chua, Tat-Seng and Zhuang, Yueting},
  booktitle={Proceedings of the IEEE/CVF International Conference on Computer Vision},
  pages={1718--1728},
  year={2025}
}

@inproceedings{li2024image,
  title={Image Semantic Segmentation via Chain-of-Thought Prompts},
  author={Li, Lei},
  booktitle={Proceedings of the IEEE/CVF Winter Conference on Applications of Computer Vision (WACV)},
  year={2024}
}

@inproceedings{li2026multiple,
  title={Multiple Human Motion Understanding},
  author={Li, Lei and Jia, Sen and Hwang, Jenq-Neng},
  booktitle={Proceedings of the AAAI Conference on Artificial Intelligence},
  volume={40},
  pages={6297--6305},
  year={2026}
}

@inproceedings{shi2026intrinsic,
  title={Intrinsic Entropy of Context Length Scaling in LLMs},
  author={Shi, Jingzhe and Ma, Qinwei and Liu, Hongyi and Zhao, Hang and Hwang, Jenq-Neng and Li, Lei},
  booktitle={The Fourteenth International Conference on Learning Representations},
  year={2026}
}

@inproceedings{li2025human,
title={Human Motion Instruction Tuning},
author={Li, Lei and Jia, Sen and Wang, Jianhao and Jiang, Zhongyu and Zhou, Feng and Dai, Ju and Zhang, Tianfang and Wu, Zongkai and Hwang, Jenq-Neng},
booktitle={Proceedings of the IEEE/CVF Conference on Computer Vision and Pattern Recognition (CVPR)},
year={2025}
}

@inproceedings{yao2025countllm,
title={{CountLLM: Towards Generalizable Repetitive Action Counting via Large Language Model}},
author={Yao, Ziyu and Cheng, Xuxin and Huang, Zhiqi and Li, Lei},
booktitle={Proceedings of the IEEE/CVF Conference on Computer Vision and Pattern Recognition (CVPR)},
year={2025}
}

@inproceedings{cai2025bayesian,
title={{Bayesian Optimization for Controlled Image Editing via LLMs}},
author={Cai, Chengkun and Liu, Haoliang and Zhao, Xu and Jiang, Zhongyu and Zhang, Tianfang and Wu, Zongkai and Lee, John and Hwang, Jenq-Neng and Li, Lei},
booktitle={Proceedings of the Annual Meeting of the Association for Computational Linguistics (ACL)},
year={2025}
}

@inproceedings{cai2025role,
title={{The Role of Deductive and Inductive Reasoning in Large Language Models}},
author={Cai, Chengkun and Zhao, Xu and Liu, Haoliang and Jiang, Zhongyu and Zhang, Tianfang and Wu, Zongkai and Hwang, Jenq-Neng and Li, Lei},
booktitle={Proceedings of the Annual Meeting of the Association for Computational Linguistics (ACL)},
year={2025}
}

@article{jia2026ram,
  title={RAM: Recover Any 3D Human Motion in-the-Wild},
  author={Jia, Sen and Zhu, Ning and Zhong, Jinqin and Zhou, Jiale and Zhang, Huaping and Hwang, Jenq-Neng and Li, Lei},
  journal={arXiv preprint arXiv:2603.19929},
  year={2026}
}

@InProceedings{Yan_2026_WACV,
    author    = {Yan, Ziyang and Shao, Yihua and Liao, Minwen and Chen, Siyu and Wang, Nan and Lin, Muyuan and Hwang, Jenq-Neng and Zhao, Hao and Remondino, Fabio and Li, Lei},
    title     = {3DSceneEditor: Controllable 3D Scene Editing with Gaussian Splatting},
    booktitle = {Proceedings of the IEEE/CVF Winter Conference on Applications of Computer Vision (WACV)},
    month     = {March},
    year      = {2026},
    pages     = {1852-1863}
}

@article{liu2024graph,
  title={Graph canvas for controllable 3d scene generation},
  author={Liu, Libin and Chen, Shen and Jia, Sen and Shi, Jingzhe and Jiang, Zhongyu and Jin, Can and Zongkai, Wu and Hwang, Jenq-Neng and Li, Lei},
  journal={arXiv preprint arXiv:2412.00091},
  year={2024}
}

@article{guan2025learning,
  title={Learning an Efficient Optimizer via Hybrid-Policy Sub-Trajectory Balance},
  author={Guan, Yunchuan and Liu, Yu and Zhou, Ke and Li, Hui and Jia, Sen and Shen, Zhiqi and Wang, Ziyang and Zhang, Xinglin and Chen, Tao and Hwang, Jenq-Neng and others},
  journal={arXiv preprint arXiv:2511.00543},
  year={2025}
}

@article{zhang2026psgs,
  title={PSGS: Text-driven Panorama Sliding Scene Generation via Gaussian Splatting},
  author={Zhang, Xin and Chen, Shen and Zhou, Jiale and Li, Lei},
  journal={arXiv preprint arXiv:2602.00463},
  year={2026}
}

@article{yan20243dsceneeditor,
  title={3dsceneeditor: Controllable 3d scene editing with gaussian splatting},
  author={Yan, Ziyang and Li, Lei and Shao, Yihua and Chen, Siyu and Wu, Zongkai and Hwang, Jenq-Neng and Zhao, Hao and Remondino, Fabio},
  journal={arXiv preprint arXiv:2412.01583},
  year={2024}
}

@inproceedings{chen2025dense,
  title={Dense point clouds matter: Dust-gs for scene reconstruction from sparse viewpoints},
  author={Chen, Shen and Zhou, Jiale and Li, Lei},
  booktitle={ICASSP 2025-2025 IEEE International Conference on Acoustics, Speech and Signal Processing (ICASSP)},
  pages={1--5},
  year={2025},
  organization={IEEE}
}

\end{document}